\documentclass[letterpaper, 10 pt, conference]{ieeeconf}
\IEEEoverridecommandlockouts
\usepackage{cite}
\usepackage{amsmath,amssymb,amsfonts}
\usepackage{algorithmic}
\usepackage{graphicx}
\usepackage{textcomp}
\usepackage{xcolor}
\usepackage{booktabs}
\usepackage{siunitx}
\usepackage{caption} 
\usepackage{multirow}
\usepackage{makecell}
\usepackage{amsmath}

\newif\ifblind
\blindfalse        

\newcommand{\best}[1]{\textbf{#1}}
\newcommand{\coolname}{\emph{FLEET}}
\def\BibTeX{{\rm B\kern-.05em{\sc i\kern-.025em b}\kern-.08em
    T\kern-.1667em\lower.7ex\hbox{E}\kern-.125emX}}
\title{FLEET: Formal Language-Grounded Scheduling for Heterogeneous Robot Teams}

\ifblind
  \author{\IEEEauthorblockN{Anonymous Authors}\\
          \IEEEauthorblockA{Affiliation withheld for double-blind review}}
\else
\author{
Corban Rivera$^{\dagger1,2}$,
Grayson Byrd$^{1,2}$,
Meghan Booker$^{1}$,
Bethany Kemp$^{1}$,
Allison Gaines$^{1}$,\\
Emma Holmes$^{1}$, 
James Uplinger$^{3}$,
Celso M de Melo$^{3}$,
David Handelman$^{1}$\\
$^{1}$JHU APL,
$^{2}$JHU,
$^{3}$DEVCOM ARL
\thanks{$\dagger$Project Lead  *Equal Contribution}
}
\fi

\begin{document}

\makeatletter
\let\@oldmaketitle\@maketitle
\renewcommand{\@maketitle}{\@oldmaketitle
\centering
\setcounter{figure}{0}
\includegraphics[width=\linewidth]{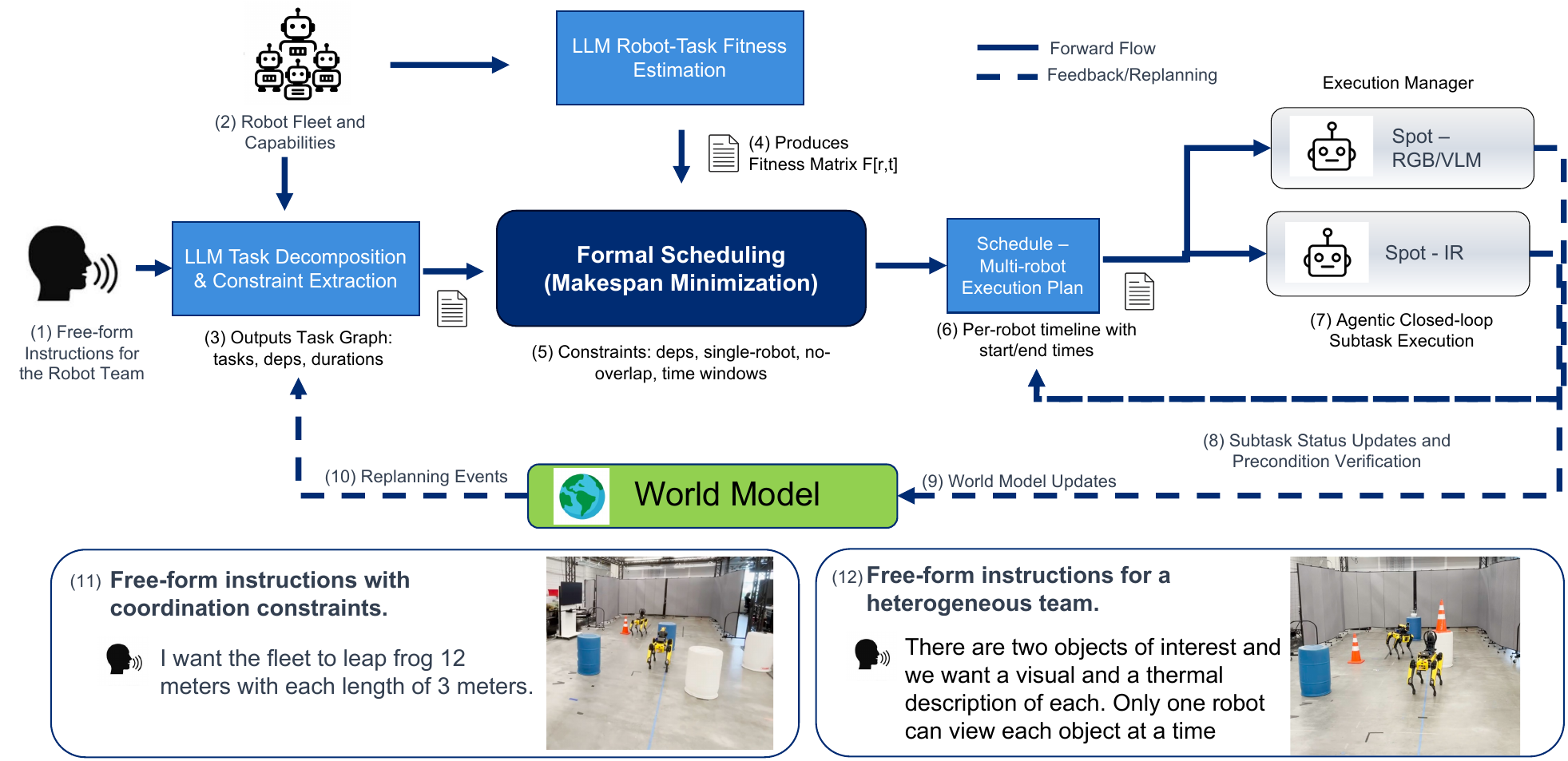}
\captionof{figure}{
\textbf{FLEET — Formal Language-grounded Execution and Efficient Teaming} is a hybrid generative–formal framework for natural-language multi-robot tasking. Free-form operator instructions (bottom examples) are ingested by an LLM that (3) decomposes the command into a task graph and constraints and (4) estimates a robot–task fitness matrix. A formal mixed-integer linear programming (MILP) scheduler solves a makespan-minimization problem under precedence, capacity, and spatial constraints to produce a multi-robot schedule (6). Robots execute the plan (7) while streaming status and perception to a World Model (8-9); deviations (delays, failures, new detections) trigger closed-loop replanning back to the LLM and scheduler (10). The architecture supports heterogeneous teams (e.g., IR and RGB/VLM Spots) and yields interpretable artifacts—task graph, fitness matrix, and schedule—that explain decisions.
}
\label{fig:splash}
\vspace{-0.98em}
}
\makeatother


\maketitle

\begin{abstract}
Coordinating heterogeneous robot teams from free-form natural-language instructions is hard: language-only planners struggle with long-horizon coordination and hallucination, while purely formal methods require closed-world models. We present \textsc{\coolname}, a hybrid decentralized framework that turns language into optimized multi-robot schedules. An LLM front-end produces (i) a task graph with durations and precedence and (ii) a capability-aware robot–task fitness matrix; a formal back-end solves a makespan-minimization problem while the underlying robots execute their free-form subtasks with agentic closed-loop control. Across multiple free-form language-guided autonomy coordination benchmarks, \textsc{\coolname} improves success over state of the art generative planners on two-agent teams across  heterogeneous tasks. Ablations show that mixed integer linear programming (MILP) primarily improves temporal structure, while LLM-derived fitness is decisive for capability-coupled tasks; together they deliver the highest overall performance. We demonstrate the translation to real world challenges with hardware trials using a pair of quadruped robots with disjoint capabilities.
\end{abstract}

\section{INTRODUCTION}
Coordinating heterogeneous robot teams in open-world environments remains a central challenge in robotics. Unlike structured factories or warehouses, homes, offices, and disaster sites are dynamic, partially observed, and underspecified at design time. A planner must translate free-form instructions into executable multi-robot strategies, respect capability and resource constraints, and adapt online to delays, perception errors, and newly discovered goals. Purely formal methods provide guarantees but assume closed-world models with carefully engineered predicates and costs; purely generative methods (e.g., LLM planners) offer semantic flexibility and rapid iteration from language, but struggle with long-horizon coordination, precedence tracking, and hallucination.

We subscribe to the recent trend that language models and formal optimization are complementary. The contributions of hybrid approaches have been primarily demonstrated on single robots.  In this work, we introduce hybrid generative and formal optimization concepts for multi-robot coordination.  LLMs excel at \emph{semantic front-end} tasks—decomposing natural language into subtasks, exposing commonsense ordering, and explaining why a robot is (or is not) suitable for a role—while a formal \emph{back-end} can guarantee feasibility and optimize the team schedule under explicit constraints. This paper presents \textsc{\coolname}, illustrated in Figure \ref{fig:splash}, a hybrid framework that embeds LLM-derived artifacts into a makespan-minimizing scheduler. Given a free-form instruction and a short profile of each robot, the system produces: (i) a task graph with durations and precedence, (ii) a robot–task fitness matrix aligned with capabilities, and (iii) a multi-robot schedule with start times and assignments. The scheduler enforces precedence, non-overlap on each robot. It runs in \emph{anytime} mode with tight caps and falls back to fast Auction allocators when needed, yielding interpretable plans that can be executed by lightweight reasoning agents.

Empirically, over multiple free-form language-guided autonomy coordination benchmarks~\cite{partnr}, our method improves success on heterogeneous team tasks over strong language-model planners, and ablations show that the combination of capability-aware fitness and formal scheduling is crucial. Hardware trials with two Boston Dynamics Spots (IR vs.\ RGB/VLM) highlight safety benefits (deconfliction in the center field) and reduced idle during cross-modal inspections.

\section{RELATED WORK}

\subsection{Formal Methods for Multi-Robot Coordination}
Classic Multi-Robot Task Allocation (MRTA)~\cite{mrta,taxonomy} includes exact assignment with Hungarian~\cite{hungarian}, market-based auctions and contracts for scalability~\cite{sold,market,auction,taxonomy}, and greedy/list-scheduling heuristics with provable bounds in simplified settings~\cite{greedy}. Beyond one-shot pairing, temporal/resource constraints are handled via MILP or constraint programming~\cite{formal}, yielding optimal or bounded-suboptimal schedules with explicit feasibility guarantees. These approaches rely on structured models and carefully tuned costs, which limits application to open-world, language-specified missions.

\subsection{Generative Planning for Single Robots}
Language-guided decision making has advanced via LLM-grounded planners and vision-language action models, e.g., SayCan/SayNav~\cite{saycan,saynav}, RT-1/RT-2 and related policy models~\cite{rt1,rt2,decision_transformer}, and tool-use agents such as ReAct and successors~\cite{react,language_planners}. Low-level control is often delegated to trajectory optimizers or learned diffusion/flow policies~\cite{diffusion1,flow_matching}. While these systems can follow open-ended instructions and improvise with tools, they typically plan for \emph{one} robot and degrade on long-horizon, tightly constrained tasks.

\subsection{LLM-based Centralized Multi-Robot Planning}
LLMs have been used to map team-level goals to subtask assignments and dialogue among agents~\cite{language_planners,coela,roco,centralized1}. SMART-LLM~\cite{smartllm} is a representative centralized planner in which a single LLM decomposes and allocates subtasks to robots. Benchmarks such as PARTNR~\cite{partnr} (built on Habitat~\cite{habitat}) expose persistent weaknesses of purely generative planners on multi-agent tasks: coordination errors, precedence violations, and poor recovery from partial failure.

\subsection{Hybrid Formal–Generative Single Robot Execution}
Recent work uses LLMs to produce symbolic structures for classical solvers—PDDL domain/goal synthesis~\cite{world_models}, temporal logic specifications~\cite{ltl,ltl2}, precondition/postcondition checking~\cite{conceptagent,pre2,pre3}, factor-graph formulations~\cite{factor_graphs}, or linear programs~\cite{milp}. Two-stage pipelines show that LLMs can propose subgoals which are then solved by a combinatorial back-end~\cite{twostep}. Our approach differs by focusing on multi-robot coordination and \emph{jointly} using (i) an LLM-derived task graph and (ii) an LLM-derived, capability-aware fitness matrix as \emph{inputs} to a makespan-minimizing scheduler. This yields feasible, optimized schedules that remain semantically aligned with the original instruction and robot capabilities, and it enables clear ablations (MILP only, fitness only) and fast fallback to auction assignment under strict latency caps.

\subsection{Contributions}
We contribute a practical and interpretable pipeline for language-guided multi-robot coordination in open-world settings:
\begin{itemize}
  \item \textbf{Hybrid planning architecture.} An LLM front-end produces a task graph and a robot–task fitness matrix; a formal back-end (MILP with anytime settings) computes a schedule that enforces precedence, non-overlap.
  \item \textbf{Allocator plug-ins and anytime operation.} The scheduler supports MILP, Auction back-ends behind a shared interface; MILP runs with strict time/gap caps and warm-starts, and the system falls back to heuristics to guarantee progress.
  \item \textbf{Execution with reasoning agents.} Each robot executes assigned subtasks via a constrained tool-use loop (navigation, perception, manipulation) and streams status to a shared world model, triggering event-based replanning when needed.
  \item \textbf{Comprehensive evaluation.} We compare against state-of-the-art language-model planners and run ablations isolating the roles of fitness and formal scheduling across multiple PARTNR benchmark categories. Hardware trials with two heterogeneous Spots validate safety (deconfliction) and efficiency (reduced idle) in cross-modal tasks.
\end{itemize}

\section{METHODS}

\begin{figure*}[th]
  \centering
  \includegraphics[width=\textwidth]{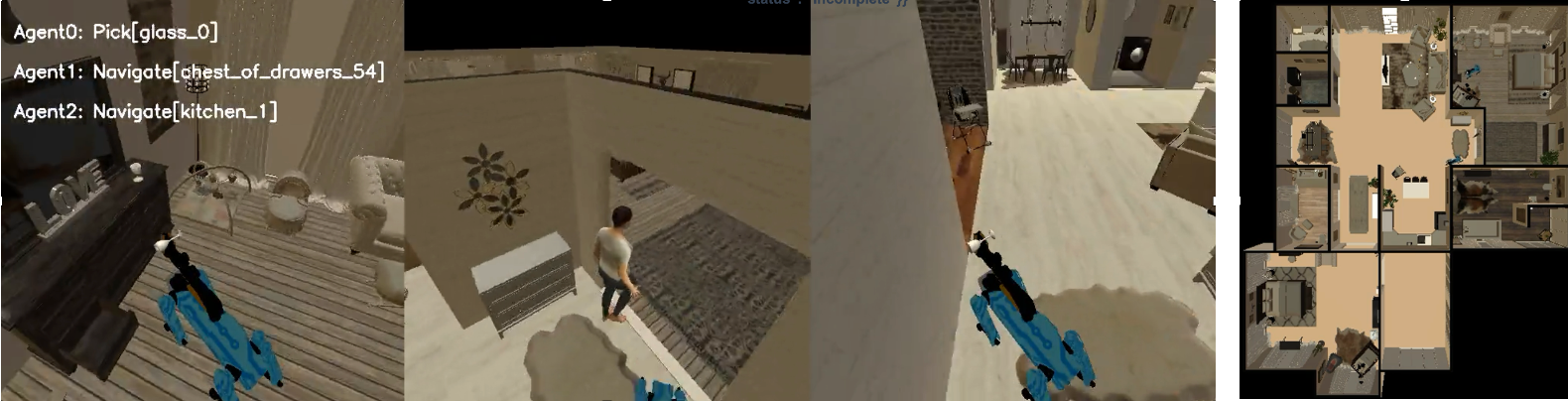}
  \caption{\textbf{PARTNR Free-form Language-guided Benchmarks} Partnr free-form language mulit-agent benchmarks builds on habitat-sim and introduces several categories of free-form language-guided tasks to be completed by one or more agents.  The categories include "Constraint Free" where the subtasks are separable and do not necessarily depend on each other, "Heterogeneous" where the agents have disjoint capabilities that must be leveraged correctly to complete the tasks, and "Temporal" where the tasks have an implied dependency structure among the subtasks. This Figure illustrates a task from the Heterogeneous task set where agents are acting on the command "Take all of the glasses from the bedroom to the kitchen and wash them". In this example, the human agent can clean and the quadruped robots can not. }
  \label{fig:partnr}
\end{figure*}

\begin{figure}[th]
  \centering
  \includegraphics[width=\columnwidth]{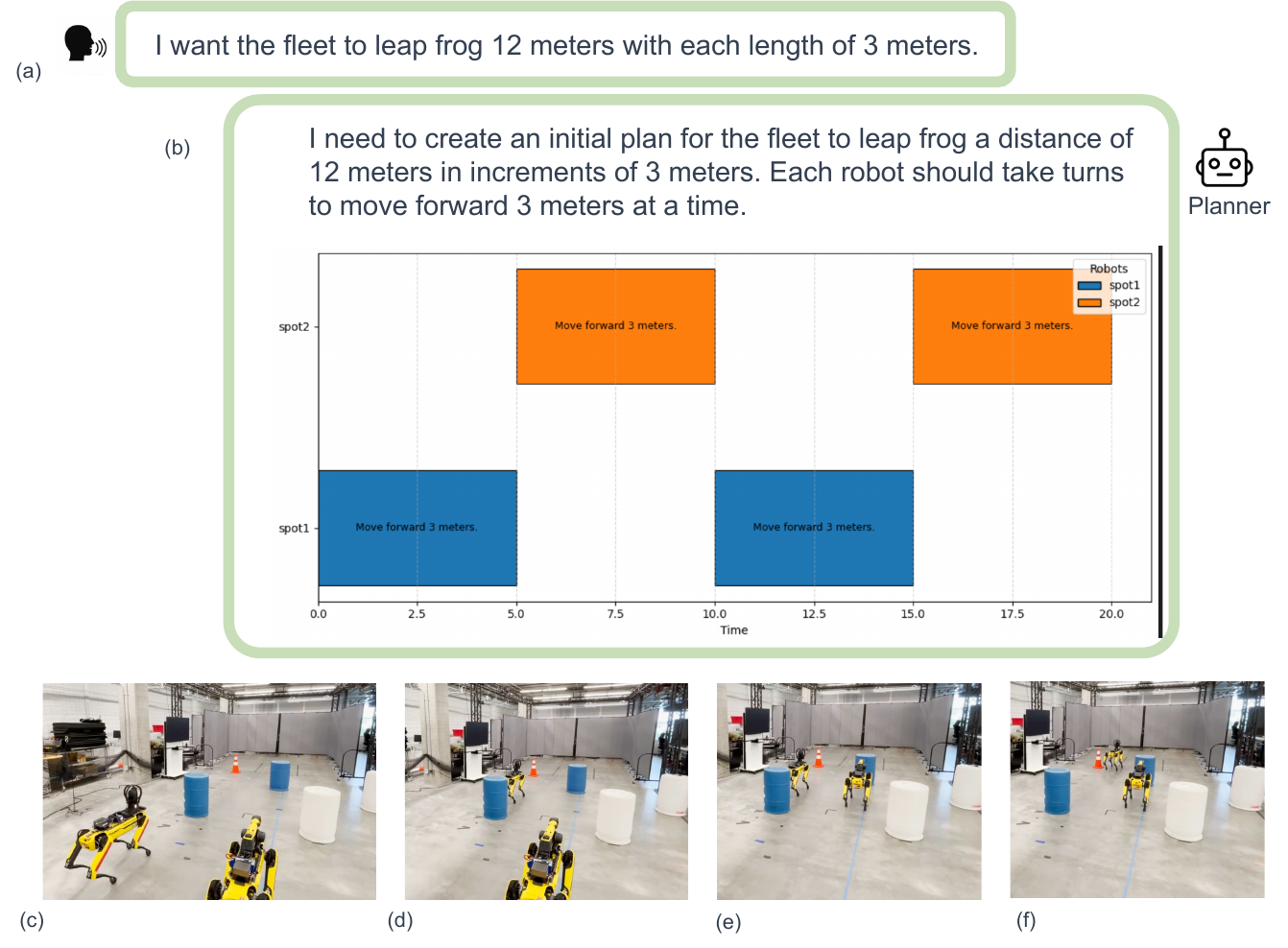}
  \caption{Hardware trial: \textbf{Maneuver with implied dependencies}
  (a) Operator instruction. (b) Planner output: schedule with 3\,m segments and
  enforced alternation. (c–f) Execution frames showing alternating advances.}
  \label{fig:leapfrog}
\end{figure}

\begin{figure*}[th]
  \centering
  \includegraphics[width=\textwidth]{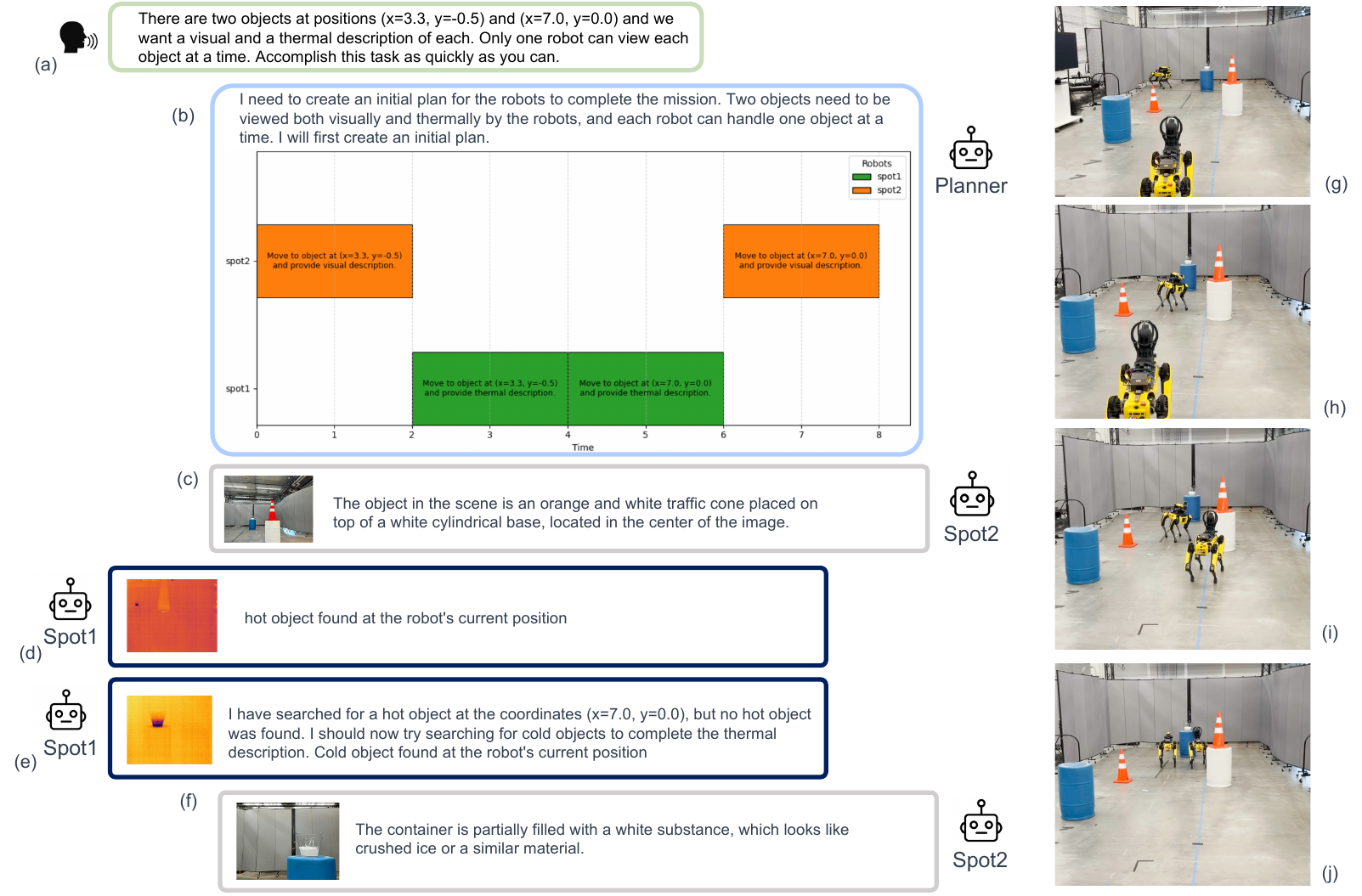}
  \caption{Hardware trial: \textbf{Cross-modal inspection.}  The environment included a heat pad under a traffic cone at point-of-interest 1 and a bucket of ice at point of interest 2.  The robot team was asked to visually and thermally characterize both points of interest with the additional constraint that only one robot could analyse a point of interest at a time.  The robots have disjoint capabilities where Spot-IR can only provide thermal analysis and Spot-RGB/VLM is the only robot that can provide visual question answering.
  (a) Natural-language instruction. (b) Planner schedule with AND-dependencies
  (visual+thermal). (c–f) robot QA responses (RGB/IR). (g–j) execution frames.
  The formal scheduler releases steps on dependency completion, reducing idle
  time and handoff latency. }
  \label{fig:crossmodal-twocol}
\end{figure*}

Our framework blends the semantic flexibility of LLMs with the coordination guarantees of formal optimization. A single pipeline converts a free-form instruction into three concrete artifacts: (i) a \emph{task graph} with durations and precedence, (ii) a \emph{robot--task fitness matrix} that captures capabilities and preferences, and (iii) a \emph{multi-robot schedule} (start times and assignments).

\subsection{Problem Setup and Artifacts}
We consider a heterogeneous team $R=\{1,\dots,n\}$ and a set of tasks $T=\{1,\dots,m\}$. Each task $j$ has duration $d_j>0$, optional precedence constraints $\text{Pred}(j)\!\subseteq\!T$, and optional time windows $[r_j,\ell_j]$. Robots have capabilities $\mathcal{C}_i$ (e.g., thermal QA, VLM QA), which induce a feasibility mask $g_{ij}\in\{0,1\}$ (task $j$ is feasible for robot $i$ iff required capabilities are in $\mathcal{C}_i$). The LLM also returns a normalized fitness score $f_{ij}\!\in\![0,1]$ indicating how suitable robot $i$ is for task $j$ (higher is better).

\paragraph{Artifacts produced by the LLM.}
Given a natural-language instruction and short robot profiles, the LLM outputs a JSON task list:
\begin{itemize}
  \item \texttt{id}, \texttt{description} (subtask string), \texttt{duration} $d_j$,
  \item \texttt{dependencies} (list of predecessors), and optional \texttt{constraints}:
         \texttt{location} (for travel accounting).
\end{itemize}

A second prompt maps robot profiles to a scalar $f_{ij}$, forming a matrix $F\!\in\![0,1]^{n\times m}$.
In practice we use few-shot, chain-of-thought style grading and min–max normalization across robots per task to reduce bias. When $F$ is unavailable, we default to uniform scores.

\subsection{Scheduling Back-end: MILP Formulation}
Let $R=\{1,\dots,n\}$ denote robots and $T=\{1,\dots,m\}$ tasks. Each task $j$ has duration $d_j>0$ and precedence edges
$E\subseteq T\times T$; $(k,j)\in E$ means $k$ must finish before $j$ starts.
Feasibility $g_{ij}\in\{0,1\}$ encodes capabilities; fitness $f_{ij}\in[0,1]$ induces a linear cost
$c_{ij}=\tfrac{1}{1+\gamma f_{ij}}+\tau\,\text{travel}_{ij}$ (set $\tau{=}0$ if travel is unused).
Pick $M\ge \sum_{j\in T} d_j$.
Decision variables: $x_{ij}\in\{0,1\}$ (assign $j$ to $i$), $s_j\ge 0$ (start time), $y^i_{jk}\in\{0,1\}$ (on robot $i$, $j$ before $k$),
$C_i\ge 0$ (robot completion), and $C_{\max}\ge 0$ (makespan).

\begin{align}
\min\ & \alpha\,C_{\max} + \beta \sum_{i\in R} C_i
      + \lambda \sum_{i\in R}\sum_{j\in T} c_{ij}\,x_{ij}
\label{eq:obj}
\end{align}
\noindent\textit{Prioritizes makespan ($\alpha$) with secondary load balance ($\beta$) and a soft preference for capability/fit and optional travel cost ($\lambda$).}

\begin{align}
\sum_{i\in R} x_{ij} = 1 \qquad &\forall j\in T \label{eq:assign}\\
x_{ij} \le g_{ij}               \qquad &\forall i\in R,\ \forall j\in T \label{eq:feas}\\
s_j \ge s_k + d_k               \qquad &\forall (k,j)\in E \label{eq:prec}
\end{align}
\noindent\textit{(\ref{eq:assign}) assigns each task to exactly one robot; (\ref{eq:feas}) enforces capability feasibility; (\ref{eq:prec}) respects all precedence edges.}

\begin{align}
s_j + d_j \;\le\;
\begin{aligned}[t]
  &\, s_k + M(1 - y_{jk}^i)\\
  &{}+ M(2 - x_{ij} - x_{ik})
\end{aligned}
\qquad &\forall i\in R,\ \forall j<k
\label{eq:noov1}
\end{align}
\noindent\textit{if $x_{ij}{=}x_{ik}{=}1$ and $y_{jk}^i{=}1$, then $j$ must finish before $k$ on robot $i$; otherwise the big-$M$ terms relax the constraint.}

\begin{align}
s_k + d_k \;\le\;
\begin{aligned}[t]
  &\, s_j + M\,y_{jk}^i\\
  &{}+ M(2 - x_{ij} - x_{ik})
\end{aligned}
\qquad &\forall i\in R,\ \forall j<k
\label{eq:noov2}
\end{align}
\noindent\textit{Symmetric branch—if $x_{ij}{=}x_{ik}{=}1$ and $y_{jk}^i{=}0$, then $k$ must finish before $j$ on robot $i$; otherwise it relaxes. Together with (\ref{eq:noov1}) this prevents overlap on the same robot.}

\begin{align}
C_i \;\ge\;
\begin{aligned}[t]
  &\, s_j + d_j\\
  &{}- M(1 - x_{ij})
\end{aligned}
\qquad &\forall i\in R,\ \forall j\in T
\label{eq:Ci}
\end{align}
\noindent\textit{Robot completion time $C_i$ lower-bounds the finish time of every task assigned to robot $i$.}

\begin{align}
C_{\max} \ge s_j + d_j \qquad &\forall j\in T.
\label{eq:Cmax}
\end{align}
\noindent\textit{Makespan $C_{\max}$ lower-bounds the finish time of every task (overall completion).}

\emph{Anytime MILP and fallbacks.} --
\coolname run CBC with a wall-clock cap and a gap stop (e.g., \texttt{timeLimit=120\,s}, \texttt{gapRel=1\%}); on timeout, the incumbent plan is used. If no incumbent exists, \coolname falls back to an \emph{Auction} allocator (same I/O schema). This ensures progress during evaluation and enables quality/latency trade-offs.

\subsection{Pluggable Allocators (Auction, MILP)}
For scalability studies we swap the MILP with:
 \textbf{Auction:} robots bid for ready tasks using costs $c_{ij}$, iterating until $\epsilon$-optimality; ties break by earliest finish time and robot ID. Dependencies gate task readiness \cite{auction}.
All allocators emit the same schedule dictionary (agent ID, start, end, metadata), which we visualize as a Gantt chart and push to the execution layer.

\subsection{Closed-loop Execution with Reasoning Agents}
The global schedule specifies \emph{who} does \emph{what} and \emph{when}. Each robot executes its assigned free-form language subtasks with \emph{ConceptAgent} \cite{conceptagent}: the LLM can call a small set of verified tools (navigation, perception, manipulation), must return JSON under a fixed schema, and logs status back to a shared world model (task state, detections, poses). \emph{Triggers} for replanning include (i) task completion, (ii) delay beyond a threshold, (iii) perception contradictions (e.g., VLM/IR mismatch), or (iv) newly discovered obstacles. On replanning, we re-score fitness only for impacted tasks and resolve with the same allocator (MILP/Auction), warm-starting from the current partial schedule to preserve stability.  Prompt formats are derived from PARTNR \cite{partnr} decentralized baselines.

\emph {Implementation details.} --
We use fixed prompt templates (few-shot for both decomposition and fitness scoring), normalize $f_{ij}$ per task, and clamp durations to positive values. For CBC we set \texttt{presolve=true}, enable cuts, and cap threads to available cores. $M$ is chosen as $\sum_{j} d_j$, and time windows are omitted if absent. When locations are present we include travel either in $c_{ij}$ (cost term) or by augmenting $d_j$ with the preceding leg’s travel time; both variants are supported by our code.

\section{RESULTS}

\subsection{Evaluation in Simulation (PARTNR)}
We evaluate on PARTNR~\cite{partnr}, illustrated in Figure \ref{fig:partnr}, which scores success on free-form, language-guided multi-robot tasks. Following its taxonomy, we report three categories: (i) \emph{constraint-free} (subtasks can be completed in any order by any agent), (ii) \emph{temporal} (explicit/implicit precedence constraints), and (iii) \emph{heterogeneous} (capability-restricted subtasks). PARTNR provides a verified evaluation function per instruction (goal propositions and constraints), enabling automatic scoring. We omit strictly spatial-relation tasks and focus on categories most aligned with capability- and order-aware scheduling.

\paragraph{Protocol and metrics}
Unless noted, we measure task \emph{success rate} (fraction completed; higher is better; binary variable). We evaluate 2- and 3-agent (when possible) teams. Prompts, seeds, and full configuration are in the supplement.   We report macro-averages across PARTNR categories; all methods share the same instruction sets and LLM back-ends.

\subsection{Comparison to State-of-the-Art Planners (Two Agents)}
In Table \ref{tab:sota}, we compare \textsc{FLEET} against a decentralized baseline from PARTNR and the centralized LLM planner SMART-LLM. Scores are averaged over runs with both \texttt{gpt-4o} and an open-weight \texttt{gpt-oss-20b}.

\begin{table}[t]
  \centering
  \caption{PARTNR-sim success rate (↑) for two agents. Best per column in \textbf{bold}.}
  \label{tab:sota-two}
  \footnotesize
  \setlength{\tabcolsep}{4pt}
  \renewcommand{\arraystretch}{1.15}
  \begin{tabular}{l
                  S[table-format=1.2]
                  S[table-format=1.2]
                  S[table-format=1.2]
                  S[table-format=1.2]}
    \toprule
    \textbf{Method} &
      {\makecell{\textbf{Constraint}\\\textbf{Free}}} &
      {\makecell{\textbf{Hetero-}\\\textbf{geneous}}} &
      {\textbf{Temporal}} &
      {\textbf{Average}} \\
    \midrule
    \textsc{FLEET} (Ours)       & \best{0.56} & \best{0.67} & 0.53 & \best{0.59} \\
    PARTNR \cite{partnr}        & 0.32          & 0.25          & 0.26 & 0.28 \\
    SMART-LLM \cite{smartllm}                   & \best{0.56} & 0.60          & \best{0.56} & 0.57 \\
    \bottomrule
  \end{tabular}
  \label{tab:sota}
\end{table}

\emph{Findings.} --
(1) \emph{Against decentralized planning.} \textsc{FLEET} improves substantially over the PARTNR decentralized baseline across all categories: +0.24 (constraint-free), +0.42 (heterogeneous), +0.27 (temporal), and +0.31 overall (0.59 vs.\ 0.28). This reflects the benefit of explicit precedence/resource constraints and capability-aware assignment.  

(2) \emph{Against SMART-LLM.} \textsc{FLEET} is comparable with SMART - LLM on constraint-free tasks (0.56 vs.\ 0.56) and temporal tasks (0.53 vs.\ 0.56), is stronger on heterogeneous tasks (0.67 vs.\ 0.60, +0.07), yielding a higher overall average (0.59 vs.\ 0.57). The heterogeneous gain is consistent with our fitness-guided allocation.  

A hybrid approach that pairs LLM-derived task graphs and fitness with a formal scheduler yields competitive constraint-free performance, stronger capability-coupled coordination, and state-of-the-art average success with two agents.

\subsection{Ablation Studies}
\label{sec:ablations}
We ablate two components of \textsc{FLEET}: the \emph{formal scheduler} (MILP) and the \emph{LLM-based robot–task fitness}. The base variant uses uniform fitness with no formal optimizer; \textsc{FLEET} (+MILP) adds the scheduler; \textsc{FLEET} (+MILP + LLM fitness) adds capability-aware fitness on top of MILP. Scores are macro-averaged across task categories.

\begin{table}[t]
  \centering
  \caption{Ablations on \textsc{FLEET}. Success rate (↑). Avg = macro-average. Best per column in \textbf{bold}.}
  \label{tab:ablate}
  \footnotesize
  \setlength{\tabcolsep}{3pt} 
  \renewcommand{\arraystretch}{1.15}
  \begin{tabular}{l l
                  S[table-format=1.2]
                  S[table-format=1.2]
                  S[table-format=1.2]
                  S[table-format=1.2]}
    \toprule
\textbf{Method} &\textbf{Ablations} & {\makecell{\textbf{Constraint}\\\textbf{Free}}} & {\makecell{\textbf{Hetero-}\\\textbf{geneous}}} & {\textbf{Temporal}} & {\textbf{Average}} \\
    \midrule
    \multirow{4}{*}{\textsc{FLEET}}
      & \makecell[l]{[+MILP\\+LLM fitness]} & \textbf{0.56} & \textbf{0.67} & \textbf{0.53} & \textbf{0.59} \\
      & \makecell[l]{[+auction\\+LLM fitness]} & \textbf{0.56} & 0.50 & 0.44 & 0.50 \\
      & [+MILP]               & 0.41          & 0.27          & 0.49          & 0.39 \\
      & [base]                 & 0.32          & 0.25          & 0.26          & 0.28 \\
    \bottomrule
  \end{tabular}
\end{table}

\emph{Findings} --
(1) \textbf{MILP mainly improves temporal structure.} Adding MILP to the base raises \emph{temporal} success from \(0.26\!\to\!0.49\) (\(+0.23\), \(\approx\!+88\%\) rel.), consistent with enforcing precedence. It also modestly helps \emph{constraint-free} tasks (\(0.32\!\to\!0.41\), \(+0.09\)).  The auction assignment strategy matched the performance of MILP for constraint free tasks, but had lower performance than MILP for heterogeneous and temporal tasks.

(2) \textbf{LLM fitness is decisive for heterogeneous tasks.} With uniform fitness, MILP barely changes \emph{heterogeneous} performance (\(0.25\!\to\!0.27\)). Injecting LLM-derived fitness with the same MILP lifts it to \(0.67\) (\(+0.40\) over MILP; \(+0.42\) over base), indicating that capability-aware scoring is the primary driver of correct assignments.  

(3) \textbf{Components are complementary.} The base averages \(0.28\); MILP alone reaches \(0.39\) (\(+0.11\), \(+39\%\) rel.), and MILP+fitness reaches \(0.59\) (\(+0.31\), \(+111\%\) rel.; \(+0.20\) over MILP). Constraint-free tasks also benefit from the full model (\(0.32\!\to\!0.56\), \(+0.24\)), reflecting better load balance and reduced idle from explicit scheduling guided by capability fit.  Formal scheduling and capability-aware fitness address complementary failure modes—temporal feasibility vs.\ role selection—and together yield the highest performance.

\subsection{Hardware Trials}
We tested \coolname using two Boston Dynamic Spot robots with complementary but disjoint sensing capabilities: Spot-IR (thermal QA) and Spot-RGB/VLM (visual QA), both with waypoint navigation and \texttt{home}. Each robot executes locally (ConceptAgent~\cite{conceptagent}); \textsc{\coolname} computes the schedule. Figure~\ref{fig:crossmodal-twocol} illustrates a task with heterogeneous and temporal constraints. In \emph{Cross-modal POIs}, two points of interest must be analyzed visually and thermally under a single-server constraint (only one robot may analyze a POI at a time). In \emph{Maneuver with implied dependencies}, the team advances in fixed segments with alternating motion. Figure \ref{fig:leapfrog} illustrates \coolname performing the trial. Compared to the PARTNR decentralized baseline, \textsc{\coolname} encodes precedence and schedules sensing to avoid mid-field conflicts; empirically this increases minimum inter-robot separation and reduces handoff latency and makespan on multi-modal tasks, while matching performance on leap-frog where structure is simple.

\section{CONCLUSIONS}
We presented \textsc{\coolname}, a hybrid framework that turns free-form language into optimized multi-robot schedules by pairing an LLM front-end (task graph + capability-aware fitness) with a formal back-end (MILP/Auction). The approach enforces precedence, non-overlap constraints and operates in an anytime mode for dependable progress. In simulation (PARTNR), \textsc{\coolname} improves success over strong language-model planners—especially on heterogeneous tasks—and ablations show that formal scheduling and LLM-derived fitness are complementary. Hardware trials with two heterogeneous Spots demonstrate practical benefits: deconflicted execution and shorter makespan on cross-modal inspections, while matching baselines on simpler maneuvers.



\section{ACKNOWLEDGEMENTS}

\ifblind
 Acknowledgements withheld for double-blind review
\else
The authors would like to thank David Patrone, Matthew Hahne, and Chris Cooke for hardware infrastructure support. This research was sponsored by the Army Research Laboratory and was accomplished under Cooperative Agreement Number W911NF-21-2-0211. The views and conclusions contained in this document are those of the authors and should not be interpreted as representing the official policies, either expressed or implied, of the Army Research Office or the U.S. Government. The U.S. Government is authorized to reproduce and distribute reprints for Government purposes notwithstanding any copyright notation herein. DISTRIBUTION A. Approved for public release; distribution unlimited.
\fi

\bibliographystyle{IEEEtran}
\bibliography{refs}

\begin{thebibliography}{10}
\providecommand{\url}[1]{#1}
\csname url@samestyle\endcsname
\providecommand{\newblock}{\relax}
\providecommand{\bibinfo}[2]{#2}
\providecommand{\BIBentrySTDinterwordspacing}{\spaceskip=0pt\relax}
\providecommand{\BIBentryALTinterwordstretchfactor}{4}
\providecommand{\BIBentryALTinterwordspacing}{\spaceskip=\fontdimen2\font plus
\BIBentryALTinterwordstretchfactor\fontdimen3\font minus \fontdimen4\font\relax}
\providecommand{\BIBforeignlanguage}[2]{{%
\expandafter\ifx\csname l@#1\endcsname\relax
\typeout{** WARNING: IEEEtran.bst: No hyphenation pattern has been}%
\typeout{** loaded for the language `#1'. Using the pattern for}%
\typeout{** the default language instead.}%
\else
\language=\csname l@#1\endcsname
\fi
#2}}
\providecommand{\BIBdecl}{\relax}
\BIBdecl

\bibitem{partnr}
M.~Chang, G.~Chhablani, A.~Clegg, M.~D. Cote, R.~Desai, M.~Hlavac, V.~Karashchuk, J.~Krantz, R.~Mottaghi, P.~Parashar \emph{et~al.}, ``Partnr: A benchmark for planning and reasoning in embodied multi-agent tasks,'' \emph{arXiv preprint arXiv:2411.00081}, 2024.

\bibitem{mrta}
A.~Khamis, A.~Hussein, and A.~Elmogy, ``Multi-robot task allocation: A review of the state-of-the-art,'' \emph{Cooperative robots and sensor networks 2015}, pp. 31--51, 2015.

\bibitem{taxonomy}
B.~P. Gerkey and M.~J. Matari{\'c}, ``A formal analysis and taxonomy of task allocation in multi-robot systems,'' \emph{The International journal of robotics research}, vol.~23, no.~9, pp. 939--954, 2004.

\bibitem{hungarian}
H.~W. Kuhn, ``The hungarian method for the assignment problem,'' \emph{Naval research logistics quarterly}, vol.~2, no. 1-2, pp. 83--97, 1955.

\bibitem{sold}
B.~P. Gerkey and M.~J. Mataric, ``Sold!: Auction methods for multirobot coordination,'' \emph{IEEE transactions on robotics and automation}, vol.~18, no.~5, pp. 758--768, 2002.

\bibitem{market}
M.~B. Dias, R.~Zlot, N.~Kalra, and A.~Stentz, ``Market-based multirobot coordination: A survey and analysis,'' \emph{Proceedings of the IEEE}, vol.~94, no.~7, pp. 1257--1270, 2006.

\bibitem{auction}
D.~P. Bertsekas, ``A distributed algorithm for the assignment problem,'' \emph{Lab. for Information and Decision Systems Working Paper, MIT}, vol.~3, 1979.

\bibitem{greedy}
R.~L. Graham, ``Bounds on multiprocessing timing anomalies,'' \emph{SIAM journal on Applied Mathematics}, vol.~17, no.~2, pp. 416--429, 1969.

\bibitem{formal}
M.~Gombolay, R.~Wilcox, and J.~Shah, ``Fast scheduling of multi-robot teams with temporospatial constraints,'' 2013.

\bibitem{saycan}
M.~Ahn, A.~Brohan, N.~Brown, Y.~Chebotar, O.~Cortes, B.~David, C.~Finn, C.~Fu, K.~Gopalakrishnan, K.~Hausman \emph{et~al.}, ``Do as i can, not as i say: Grounding language in robotic affordances,'' \emph{arXiv preprint arXiv:2204.01691}, 2022.

\bibitem{saynav}
A.~Rajvanshi, K.~Sikka, X.~Lin, B.~Lee, H.-P. Chiu, and A.~Velasquez, ``Saynav: Grounding large language models for dynamic planning to navigation in new environments,'' in \emph{Proceedings of the International Conference on Automated Planning and Scheduling}, vol.~34, 2024, pp. 464--474.

\bibitem{rt1}
A.~Brohan, N.~Brown, J.~Carbajal, Y.~Chebotar, J.~Dabis, C.~Finn, K.~Gopalakrishnan, K.~Hausman, A.~Herzog, J.~Hsu \emph{et~al.}, ``Rt-1: Robotics transformer for real-world control at scale,'' \emph{arXiv preprint arXiv:2212.06817}, 2022.

\bibitem{rt2}
A.~Brohan, N.~Brown, J.~Carbajal, Y.~Chebotar, X.~Chen, K.~Choromanski, T.~Ding, D.~Driess, A.~Dubey, C.~Finn \emph{et~al.}, ``Rt-2: Vision-language-action models transfer web knowledge to robotic control,'' \emph{arXiv preprint arXiv:2307.15818}, 2023.

\bibitem{decision_transformer}
L.~Chen, K.~Lu, A.~Rajeswaran, K.~Lee, A.~Grover, M.~Laskin, P.~Abbeel, A.~Srinivas, and I.~Mordatch, ``Decision transformer: Reinforcement learning via sequence modeling,'' \emph{Advances in neural information processing systems}, vol.~34, pp. 15\,084--15\,097, 2021.

\bibitem{react}
S.~Yao, J.~Zhao, D.~Yu, N.~Du, I.~Shafran, K.~Narasimhan, and Y.~Cao, ``React: Synergizing reasoning and acting in language models,'' \emph{arXiv preprint arXiv:2210.03629}, 2022.

\bibitem{language_planners}
W.~Huang, P.~Abbeel, D.~Pathak, and I.~Mordatch, ``Language models as zero-shot planners: Extracting actionable knowledge for embodied agents,'' in \emph{International conference on machine learning}.\hskip 1em plus 0.5em minus 0.4em\relax PMLR, 2022, pp. 9118--9147.

\bibitem{diffusion1}
C.~Chi, Z.~Xu, S.~Feng, E.~Cousineau, Y.~Du, B.~Burchfiel, R.~Tedrake, and S.~Song, ``Diffusion policy: Visuomotor policy learning via action diffusion,'' \emph{The International Journal of Robotics Research}, p. 02783649241273668, 2023.

\bibitem{flow_matching}
K.~Nguyen, A.~T. Le, T.~Pham, M.~Huber, J.~Peters, and M.~N. Vu, ``Flowmp: Learning motion fields for robot planning with conditional flow matching,'' \emph{arXiv preprint arXiv:2503.06135}, 2025.

\bibitem{coela}
H.~Zhang, W.~Du, J.~Shan, Q.~Zhou, Y.~Du, J.~B. Tenenbaum, T.~Shu, and C.~Gan, ``Building cooperative embodied agents modularly with large language models,'' \emph{arXiv preprint arXiv:2307.02485}, 2023.

\bibitem{roco}
Z.~Mandi, S.~Jain, and S.~Song, ``Roco: Dialectic multi-robot collaboration with large language models,'' in \emph{2024 IEEE International Conference on Robotics and Automation (ICRA)}.\hskip 1em plus 0.5em minus 0.4em\relax IEEE, 2024, pp. 286--299.

\bibitem{centralized1}
Y.~Chen, J.~Arkin, Y.~Zhang, N.~Roy, and C.~Fan, ``Scalable multi-robot collaboration with large language models: Centralized or decentralized systems?'' in \emph{2024 IEEE International Conference on Robotics and Automation (ICRA)}.\hskip 1em plus 0.5em minus 0.4em\relax IEEE, 2024, pp. 4311--4317.

\bibitem{smartllm}
S.~S. Kannan, V.~L. Venkatesh, and B.-C. Min, ``Smart-llm: Smart multi-agent robot task planning using large language models,'' in \emph{2024 IEEE/RSJ International Conference on Intelligent Robots and Systems (IROS)}.\hskip 1em plus 0.5em minus 0.4em\relax IEEE, 2024, pp. 12\,140--12\,147.

\bibitem{habitat}
X.~Puig, E.~Undersander, A.~Szot, M.~D. Cote, T.-Y. Yang, R.~Partsey, R.~Desai, A.~W. Clegg, M.~Hlavac, S.~Y. Min \emph{et~al.}, ``Habitat 3.0: A co-habitat for humans, avatars and robots,'' \emph{arXiv preprint arXiv:2310.13724}, 2023.

\bibitem{world_models}
J.~Xiang, T.~Tao, Y.~Gu, T.~Shu, Z.~Wang, Z.~Yang, and Z.~Hu, ``Language models meet world models: Embodied experiences enhance language models,'' \emph{Advances in neural information processing systems}, vol.~36, pp. 75\,392--75\,412, 2023.

\bibitem{ltl}
Z.~Wei, X.~Luo, and C.~Liu, ``Hierarchical temporal logic task and motion planning for multi-robot systems,'' \emph{arXiv preprint arXiv:2504.18899}, 2025.

\bibitem{ltl2}
X.~Lin and R.~Tron, ``Adaptive bi-level multi-robot task allocation and learning under uncertainty with temporal logic constraints,'' \emph{arXiv preprint arXiv:2502.10062}, 2025.

\bibitem{conceptagent}
C.~Rivera, G.~Byrd, W.~Paul, T.~Feldman, M.~Booker, E.~Holmes, D.~Handelman, B.~Kemp, A.~Badger, A.~Schmidt \emph{et~al.}, ``Conceptagent: Llm-driven precondition grounding and tree search for robust task planning and execution,'' 2025.

\bibitem{pre2}
S.~Mukherjee, C.~Paxton, A.~Mousavian, A.~Fishman, M.~Likhachev, and D.~Fox, ``Reactive long horizon task execution via visual skill and precondition models,'' in \emph{2021 IEEE/RSJ International Conference on Intelligent Robots and Systems (IROS)}.\hskip 1em plus 0.5em minus 0.4em\relax IEEE, 2021, pp. 5717--5724.

\bibitem{pre3}
S.~S. Raman, V.~Cohen, I.~Idrees, E.~Rosen, R.~Mooney, S.~Tellex, and D.~Paulius, ``Cape: Corrective actions from precondition errors using large language models,'' in \emph{2024 IEEE International Conference on Robotics and Automation (ICRA)}.\hskip 1em plus 0.5em minus 0.4em\relax IEEE, 2024, pp. 14\,070--14\,077.

\bibitem{factor_graphs}
U.~A. Mishra, Y.~Chen, and D.~Xu, ``Generative factor chaining: Coordinated manipulation with diffusion-based factor graph,'' in \emph{ICRA 2024 Workshop $\{$$\backslash$textemdash$\}$ Back to the Future: Robot Learning Going Probabilistic}, 2024.

\bibitem{milp}
M.~Peng, Z.~Chen, J.~Yang, J.~Huang, Z.~Shi, Q.~Liu, X.~Li, and L.~Gao, ``Automatic milp model construction for multi-robot task allocation and scheduling based on large language models,'' \emph{arXiv preprint arXiv:2503.13813}, 2025.

\bibitem{twostep}
D.~Bai, I.~Singh, D.~Traum, and J.~Thomason, ``Twostep: Multi-agent task planning using classical planners and large language models,'' \emph{arXiv preprint arXiv:2403.17246}, 2024.

\end{thebibliography}

\end{document}